# AS-Net: Fast Photoacoustic Reconstruction with Multi-feature Fusion from Sparse Data

Mengjie Guo[1,†], Hengrong Lan[1,2,3,†], Changchun Yang[1], and Fei Gao[1,*]

[1] *Hybrid Imaging System Laboratory, Shanghai Engineering Research Center of Intelligent Vision and Imaging, School of Information Science and Technology, ShanghaiTech University, Shanghai 201210, China*

[2] *Chinese Academy of Sciences, Shanghai Institute of Microsystem and Information Technology, Shanghai 200050, China*

[3] *University of Chinese Academy of Sciences, Beijing 100049, China*

[†] *equal contribution*

[*] *gaofei@shanghaitech.edu.cn*

## 1. Abstract

Photoacoustic (PA) imaging is a biomedical imaging modality capable of acquiring high-contrast images of optical absorption at depths much greater than traditional optical imaging techniques. However, practical instrumentation and geometry limit the number of available acoustic sensors surrounding the imaging target, which results in the sparsity of sensor data. Conventional PA image reconstruction methods give severe artifacts when they are applied directly to the sparse PA data. In this paper, we firstly propose to employ a novel signal processing method to make sparse PA raw data more suitable for the neural network, concurrently speeding up image reconstruction. Then we propose Attention Steered Network (AS-Net) for PA reconstruction with multi-feature fusion. AS-Net is validated on different datasets, including simulated photoacoustic data from fundus vasculature phantoms and experimental data from in vivo fish and mice. Notably, the method is also able to eliminate some artifacts present in the ground truth for in vivo data. Results demonstrated that our method provides superior reconstructions at a faster speed.

## 2. Introduction

Biomedical imaging provides a comprehensive illustration of human body in multi-scale and multi-contrast for both clinical application and scientific research. A relatively new imaging modality, photoacoustic tomography (PAT), combines the high contrast of optical imaging and high penetration of ultrasonic (US) imaging 1-3. The high potentials in quantifying endogenous chromophores or exogenous contrast agents have resulted in rapidly growing for PAT, which provides important physiological parameters, such as the oxygen saturation (sO2) of hemoglobin and the metabolic rate of oxygen 4-7, compared with pure US imaging. PAT excites photoacoustic (PA) signals with pulsed laser, and detects PA signals by ultrasound transducer. However, the artifacts could disturb the quality of PA image if the procedure of data acquisition does not satisfy the temporal or spatial Nyquist criterion with an ill-posed condition. Many works have been presented to ameliorate this problem [8-11].

Currently, deep learning (DL) inspires a new area of image reconstruction to resolve the ill-posed inverse problem [12-14]. There are two schemes for PAT image reconstruction with DL: iterative reconstruction and non-iterative reconstruction. Iterative DL approach integrates DL model into traditional optimization methods to reconstruct PA image as an optimization problem. DL plays different roles at different stage of the optimization procedure. In [15,16], the authors used a model to regularize the target of optimization, which exempts the trouble adjustment of parameter. Furthermore, a part of optimization can be learnt instead of solving whole procedure. For example, Andreas Hauptmann et al. used a simple CNN to take the gradient of the data consistency term and interactive result as inputs, inspired from proximal gradient descent [17]. However, this scheme suffers huge computational cost since they have to compute forward or adjoint model alternatingly.

On the other hand, non-iterative DL approach can be further divided into three types: PA signals enhancement (signal to signal), conversion from signal to image, PA image enhancement (image to image). For pre-processing of PA signals, DL is used to improve the signal-to-noise ratio (SNR) of PA signals and recover the full bandwidth [18,19]. For second category (PA signal to image), DL directly learns the physical model from PA signal to PA image. For instance, Derek Allman et al. used CNN to distinguish the point-like source and reflection artifact from raw PA signals [20]; Tong Tong et al. proposed FPnet that takes the PA signals and the time derivative of PA signals as inputs to form PA image [21]; Hengrong Lan et al. used multi-frequencies PA data as input to reconstruct PA image [22]. These methods have to use full connection or large convolution kernel to process the asymmetric size of PA signals' matrix. Some works take partially reconstructed data as input, which reduced the difficulty of model learning [23]. The third category (image to image) used DL to process imperfect PA image to generate enhanced PA image. For example, in [24], the authors used a generative adversarial network (GAN) to remove the artifacts caused by limited-view detectors; Stephan Antholzer et al. [25] used CNN to recover the polluted PA image; Tong Lu et al. proposed LV-GAN to enhance sparse-view result in PAT [26]. Furthermore, a combination of multi-schemes can also improve the quality of reconstructed image. In [27,28], we successively proposed Ki-GAN and Y-Net to combine two schemes, which retain raw data and texture information. However, both of them do not further exploit the performance in sparse data.

To further exploit the performance of DL in sparse data, in this paper, we propose a novel Attention Steered Network (AS-Net) with multi-feature fusion to combine sparse raw PA data and rough PA image as inputs, which introduces multi-attention to infuse different features. For asymmetric size of PA signals' matrix, a folded transformation is proposed to obtain symmetric dimension, instead of using full connection or large convolution kernel, which can reduce the parameters of model significantly. Three datasets are used to validate our method. In addition, we release a group of datasets to help other researchers develop their DL algorithms.

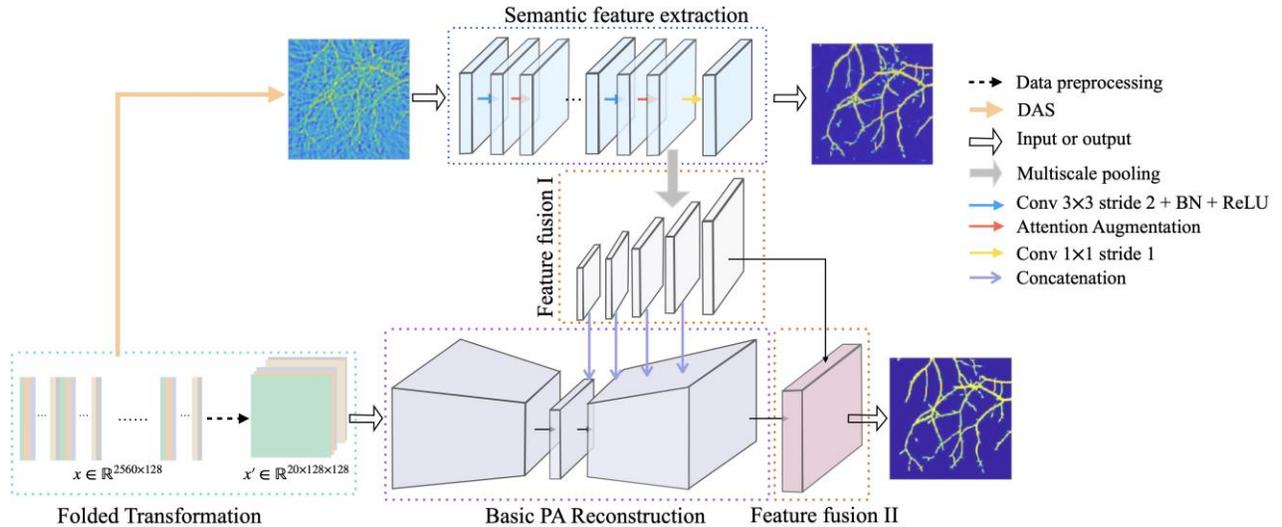

Fig. 1. Illustration of our reconstruction framework, which includes PA raw data preprocessing and AS-Net reconstruction network. Firstly, 2-D PA raw data is transformed into a 3-D square matrix by Folded Transformation (FT). Note that the input data can use arbitrary channel and length (channel can be held by zero-padding). Then our AS-Net produces the multi-feature fusion base on the attention mechanism for PA reconstruction. Our ASKF-Net architecture consists of a basic PA reconstruction (BPR) module, semantic feature extraction (SFE) module, and feature fusion (FF) module. BPR module is a modified Auto-Encoder architecture used to reconstruct images from the PA signal, while the SFE module aims to extract semantic features from the DAS image. FF module is used to fuse the semantic feature into the output of the BPR module and generate the final reconstructed image.

The main contributions of this work are summarized as follows:

We propose a signal processing method to transform the raw PA signals into a square matrix, making the input more suitable for the neural network and accelerating the reconstruction.

To reconstruct high-quality images from sparse data, we propose AS-Net, which includes a basic PA reconstruction (BPR) module, a semantic feature extraction (SFE) module, and a feature fusion (FF) module.

We apply the proposed method in different datasets, including simulated datasets, *in vivo* fish imaging datasets, and *in vivo* mouse imaging datasets. Results show that the proposed method outperforms other state-of-the-art methods on all these different datasets.

# 3. Method

In this section, we first propose a novel method for PA signal data preprocessing. We then provide details for the proposed AS-Net, which is comprised of three main components: the PA signal feature reconstruction module, the semantic feature extraction module, and the feature fusion module.

## 3.1 Folded Transformation for PA Signal

For raw PA signals, usually there are two dimensions, which represent time-domain and space-domain, respectively. PA signal in time-domain is very dense compared with sparse sensor number in space-domain, making PA signals represented by a long

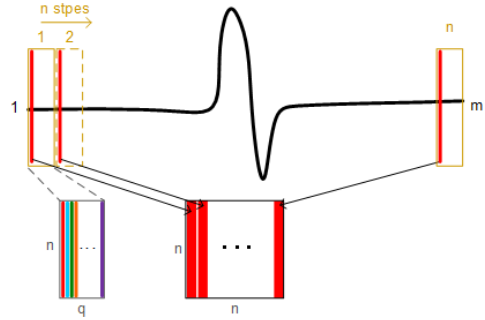
Fig.2. The illustration of folded transformation.

rectangular matrix. On the other hand, the expected reconstructed image is a square matrix. In the previous reconstruction methods with deep learning, it is common to utilize a long convolution kernel (e.g., KI-GAN [27] employs convolution kernel size 20×3 and 5×3) or full-connection to transform the long input signals into square images. An obvious drawback of using a long convolution kernel is that it will produce a large number of parameters. To address the issue, we propose a novel PA signal preprocessing method, folded transformation, which can transform the long PA signal into a square matrix. It is worth mentioning that there is no loss of information after the folded transformation because the process is reversible. The reconstruction network with the transformed signal as input can use a small convolution kernel to capture all the information in the time domain. The following is a detailed explanation of the folded transformation.

Suppose the original PA signal as $s \in \mathbb{R}^{m \times n}, m \gg n$, where $m$ represents the time-domain PA signals, $n$ represents the number of sensors in space-domain. The original signal is sampled uniformly in the time dimension at sampling frequency $q$, where

$$q = m/n \qquad (1)$$

If $q$ is not an integer, we need zero-fill $s$ to $(\lceil m/n \rceil \times n) \times n$. Therefore, we obtain $q$ matrices with size $n \times n$, where each $n \times n$ matrix can be regarded as an ultra-sparse sampling of the original PA signal. Also, the information contained between these square matrices is complementary. Namely, the original PA signal should be down-sampled at $q$ positions, and the down-sampled rate is 1:$q$, as shown in Fig. 2. Finally, we concatenate the $q$ matrices into $s'$, where $s' \in \mathbb{R}^{q \times n \times n}$, as the reconstruction network's input. For the space dimension (i.e., n), we maintain 128-channel dimensions by zero-padding, which is appropriate for different cases of sparse data. After that, we can standardize any PA signals with different sparseness and lengths. In Fig. 2, we take a PA signal received by one detector as an example to illustrate the folded transformation further.

## 3.2 Attention Steered Network (AS-Net)

*1) Overview of Network*

The architecture of the proposed AS-Net is shown in Fig. 1. As can be observed, the transformed PA signal $x'$ is fed to the BPR module, which can obtain the high-level PA features from the PA signals. On the other hand, we utilize the conventional and straightforward DAS algorithm to obtain the initial reconstruction image, which is used to extract the semantic features by the SFE module. Then the FF I module is proposed to merge the semantic features into decoders of the BPR module. Finally, the FF II module further fuses the feature maps made from BPR and FF I modules, and outputs the final PA reconstruction image.

*2) Basic PA Reconstruction Module*

In the basic PA reconstruction module, the transformed PA signal is first encoded into the high-dimensional features of the image, then decodes the feature maps into a PA image. Specifically, we modify the auto encoder-decoder network as the main body of the Basic PA reconstruction module, which consists of an encoding block, a bottom block and a decoding block.

The encoding block consists of four down-sampling blocks and two self-attention blocks, which integrate and extract the PA signal's corresponding object features. One down-sampling block is constructed by two 3×3 convolution layers.

In the bottom block, we utilize atrous residual inception [29] module, which contains four branches with atrous convolution and a residual connection. Four branches include atrous convolutions with 3×3 kernel size and [1, 3, (1, 3), (1, 3, 5)] atrous rate, respectively. The atrous residual inception module can extract semantic features of various sizes, which benefit from reconstructing PA signal images.

The decoding block is constructed by four transposed convolution layers, which can learn a self-adaptive mapping to restore features with more detailed information. The decoding block is utilized to upsample the feature maps from the bottom block and generate the initial reconstructed image.

*3) Semantic Feature Extraction Module*

The DAS reconstruction algorithm is a fast and straightforward PA image reconstruction algorithm, but DAS results are of low quality and contain many artifacts. Considering that DAS images can be obtained quickly and easily, they can provide semantic features, which can be fused into the BPR module for better reconstruction. Next, we mainly introduce the process of extracting semantic information from DAS images.

Herein, a novel DAS low-level semantic feature extractor (SFE) module is proposed, which mainly consists of four residual blocks with an attention mechanism, here referring to global context (GC) modeling operation. A bottleneck layer is employed for each residual block, which is constructed by 1×1, 3×3, and 1×1 convolution. The 1×1 layer is responsible for reducing and then increasing channels of feature maps with a bit of computational cost [30]. Furthermore, a global context block is inserted into the residual block to capture the correlation between pixels in space and extract global semantic information. Specifically, the global context feature is modeled as a weighted average of the feature map at all positions, and then added into the feature map at each position, as shown in equation (2). Finally, a skip connection is utilized to solve the gradient dispersion. The illustration of GC operation is shown in Fig. 3, and the SFE module is constructed as shown in Fig. 4.

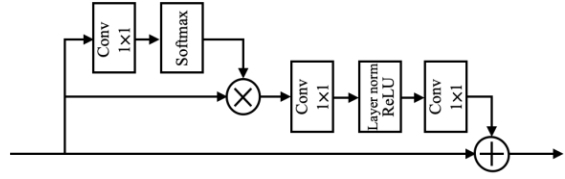

Fig. 3. The illustrations of global context (GC) block

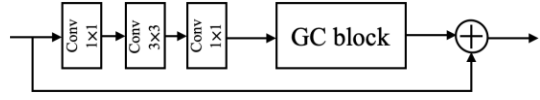

Fig. 4. The illustrations of attention augmentation of SFE module

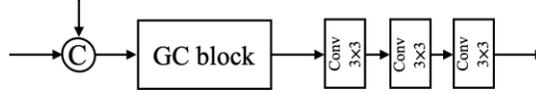

Fig. 5. The illustrations of feature fusion II

$$z_i = x_i + W_v \sum_{j=1}^{N_p} \frac{\exp(W_k x_j)}{\sum_{m=1}^{N_p} \exp(W_k x_m)} x_j \tag{5}$$

In order to ensure the validity of the semantic features extracted by the SFE module, the output of this module is supervised by an auxiliary loss function, i.e., a robust smooth $L_1$ loss function [29].

$$L_{aux} = smoothL1(y - y_d) \tag{3}$$

$$smoothL_1(x) = \begin{cases} 0.5x^2 & \text{if } |x| < 1 \\ |x| - 0.5 & \text{otherwise} \end{cases} \tag{4}$$

where $y$ denotes ground-truth image. $y_d$ denotes the output of SFE module.

*4) Feature Fusion Module*

In the previous chapter, we obtained the semantic features from DAS images. As for how to integrate the semantic features into our basic reconstruction network efficiently, the following session will give a detailed introduction.

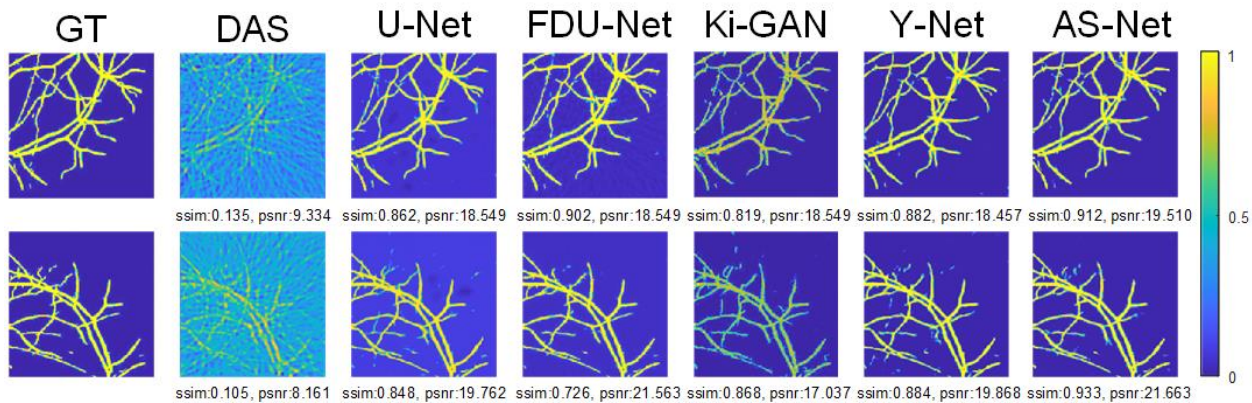

Fig. 6. Sample results of simulated fundus vessel PA image reconstruction. From left to right: ground-truth, DAS, U-Net, FDU-Net, Ki-GAN, Y-Net, Ours.

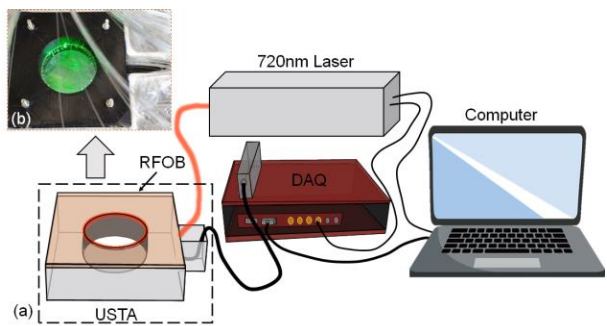

Fig. 7. (a) Schematic of the PACT system. (b) The photograph of the black box region in (a). RFOB: ring-shaped fiber optics bundle; USTA: ultrasonic transducer array; DAQ: data acquisition system.

The feature fusion module consists of two parts, FF I and FF II. FF I module is composed of five pooling layers of different sizes. The FF I transforms the output of the SFE module into different sizes by pooling, and then concatenates the features with each feature map of the bottom and decoder in the BPR module, respectively. The semantic features of the SFE module can represent different level's features with the process of pooling.

The FF II module mainly focuses on integrating reconstruction-related features further and enhancing the final reconstruction image's quality. The global context modeling operation mentioned in the last section is employed again in the FF II module. Specifically, the FF II module first concatenates the output feature map of the BPR module with the semantic features after pooling. It then inputs the fused feature to the previously defined GC block, followed by three 3×3 convolutional layers. Finally, the final reconstructed image is outputted.

After feature fusion, the network outputs the reconstructed PA image. Herein, we employ the smooth $L_1$ loss as a constraint for training the reconstruction network:

$$L_{recon} = smoothL1(y - y_r) \qquad (5)$$

where $y$ denotes ground-truth image, $y_r$ denotes the output image reconstructed by the proposed network. Finally, the overall

loss function for the proposed architecture can be defined as:

$$L_{total} = \lambda_r L_{recon} + \lambda_a L_{aux} \tag{6}$$

where $\lambda_r$ and $\lambda_a$ are the weights that balance the reconstruction loss and auxiliary loss during training. Here we set $\lambda_a = 1$, $\lambda_r = 0.2$. The total loss is optimized by Adam Optimizer to stabilize the training process and ensure that the reconstructed image is visually reasonable.

## 4. Experiments

4.1 Datasets

To evaluate the performance of the proposed method, we conducted comprehensive experiments on three datasets, which are Simulated Vessel Data, *in vivo* Fish Data, and *in vivo* Mice data. The detailed introduction to these data sets is as follows.

1) The Simulated Vessel Dataset.

We use the publicly available datasets of fundus oculi [30] as the photoacoustic initial pressure distribution. The k-wave toolbox [31] in MATLAB is used to generate the raw data, where a virtual US array with 32 transducer elements and 18 mm radius is used. The pixel size of the initial pressure map is 128 ×128. The center frequency of the ultrasound transducer is set as 5 MHz with 80% fractional bandwidth, and the propagation velocity of ultrasound is 1500 m/s.

The raw data have 2560 points with a 40 MHz sampling rate. The Simulated Vessel Dataset contains 4000 samples, of which 3600 samples are selected randomly for training and 400 for testing.

2) The Fish Data *in vivo*

The proposed approach was also tested experimentally by tomographic PA imaging of fish *in vivo*. We used a dedicated whole-body small animal imaging scanner based on cross-sectional tomographic geometry. PA signal detection is performed by a ring-shaped transducer consisting of 128 individual elements with a 30 mm radius (Central frequency: 2.5 MHz, Doppler Inc.). The PACT system is employed to record PA signals for image reconstruction. As Fig. 7 shows, a pulsed laser (720 nm wavelength, 10 Hz repetition rate) is used to illuminate the object guided by a fiber optic bundle, which is evenly separated as a circle over the transducer shown in Fig. 7(b). The data sampling rate of our system is 40 MSa/s. The region of image reconstruction is 20mm×20mm. We next acquired whole fish images with the newly developed small animal imaging scanner. The fishes were placed in a water tank. With the fishes swimming, the scanner can image fishes at different positions. The 32 channels' sparse data are selected from 128 channels' data evenly.

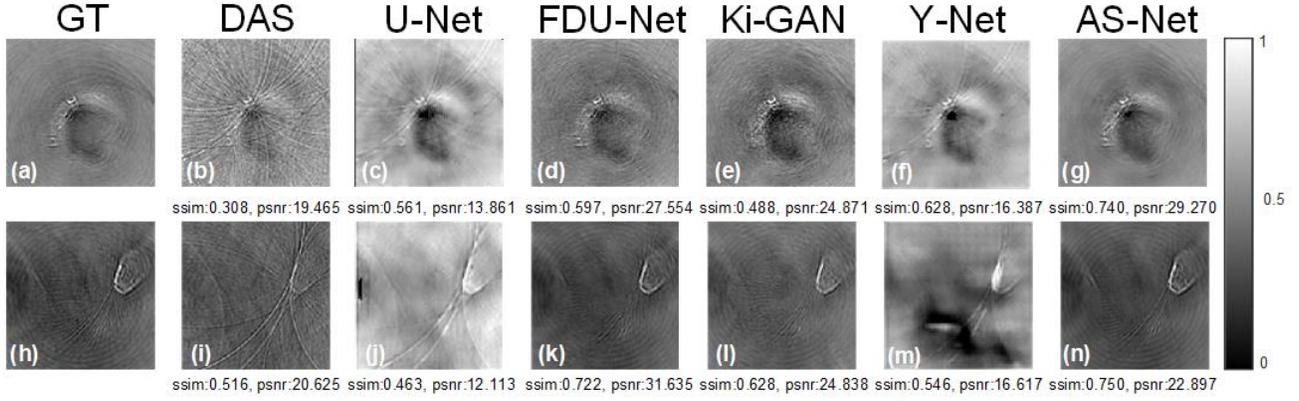

Fig. 8. Sample results of mice and fish PA image. From right to left: ground-truth, DAS, U-Net, FDU-Net, Ki-GAN, Y-Net, Ours. (a)-(g): mice result; (h)-(n): fish result.

The Fish data contains 1944 samples, and we pick 1744 for training, 200 remaining for testing, which are available at https://ieee-dataport.org/documents/his-ring-fish. The size of the PA signals is 1500×32 (12×128×128 after folded transformation). The initial reconstructed image and the ground truth's size are all 128×128.

*3) The Mice Data in vivo*

We also tested the proposed method on *in vivo* mice datasets. We acquired abdomen images from living mice using the same experimental setup. The mice data contains 1216 samples, and we pick 1046 for training, 170 remaining for testing.

The ground truth is that the DAS reconstructed images with all 128 transducer elements in the training stage. The original photoacoustic signals from the 32 probes are used as the primary input of our method, and the DAS reconstructed images of the 32 probes are used as the auxiliary input.

4.2 Evaluation Metrics

To evaluate the performance quantitatively, two metrics are adopted in this study:

*1) Structural Similarity Index Measurement (SSIM)*

SSIM evaluates the structural similarity between the reconstructed PA image and ground truth, which is more in line with the evaluation criteria of the human visual system. The SSIM can be calculated as [32]:

$$SSIM = \frac{(2\mu_y \mu_G + c_1)(2\sigma_{yG} + c_2)}{(\mu_y^2 + \mu_G^2 + c_1)(\sigma_y^2 + \sigma_G^2 + c_2)} \quad (7)$$

where $\mu_y$, $\mu_G$, $\sigma_y$ and $\sigma_G$ are the means and variances of the reconstructed image y and the ground-truth G, and $\sigma_{yG}$ is the covariance of y and G. The positive constants $c_1$ and $c_2$ are used to avoid a null denominator, where $c_1 = 0.01^2$ and $c_2 = 0.03^2$.

TABLE I
OBJECTIVE PERFORMANCE COMPARISON WITH DIFFERENT METHODS. (MEAN±STD)

| Dataset | Simulated vessel | | In-vivo Fish | | In-vivo Mice | |
|---|---|---|---|---|---|---|
| | SSIM | PSNR | SSIM | PSNR | SSIM | PSNR |
| DAS | 0.1134±0.0133 | 8.6464±0.4595 | 0.3808±0.0470 | 20.9886±0.5015 | 0.3372±0.0710 | 18.9103±2.6849 |
| U-Net | 0.8491±0.0324 | 18.3062±1.2489 | 0.4710±0.0491 | 10.3291±1.7467 | 0.5537±0.0203 | 13.9342±3.1904 |
| Y-Net | 0.8524±0.0296 | 18.3211±1.2029 | 0.5004±0.0440 | 15.5608±2.4923 | 0.6205±0.0218 | 15.5608±2.4923 |
| KI-GAN | 0.7809±0.0355 | 16.3514±0.9698 | 0.6148±0.0133 | 23.3587±1.2937 | 0.4776±0.0448 | 23.3715±1.9677 |
| FDU-Net | 0.7532±0.1569 | **19.9691±1.2172** | 0.7286±0.0587 | **29.6302±2.5573** | 0.4894±0.1622 | 22.1865±5.7565 |
| **AS-Net** | **0.8953±0.0311** | 19.5235±1.7286 | **0.7595±0.0228** | 23.5773±4.6365 | **0.7171±0.0225** | **23.2570±5.1540** |

TABLE II
TIME TO TEST AN EXAMPLE(SECONDS).

| | Model size (#params) | Model size (M) | GFLOPs (G) | Time (seconds) |
|---|---|---|---|---|
| AS-Net w/o FT | 27,923,945 | 27.923 | 90.725 | 0.143 |
| AS-Net | 10,631,049 | 10.630 | 60.396 | 0.113 |

*2) Peak Signal to Noise Ratio (PSNR)*

PSNR calculates the difference between corresponding pixels as

$$PSNR = 10\log_{10} \frac{\max^2(y,G)}{\frac{1}{N}\sum \| y-G \|_2^2} \quad (8)$$

where N is the total number of voxels in each image, and $\max^2(y,G)$ is the maximal intensity value of *y* and *G*.

Note that a higher SSIM value and higher PSNR value indicate higher quality in the reconstructed image.

4.3 Implementation Details

The proposed method was implemented in PyTorch 33 and run on a workstation equipped with four NVIDIA GPUs (GeForce TITAN RTX). The AS-Net was trained with 600 epochs and the following settings: the batch size of 16, Adam optimization with the initial learning rate of 0.005, and the learning rate linearly decay by 0.2 times for every 50 epochs. Finally, the source code is available at https://github.com/MeggieGuo/AS-Net-for-PA-reconstruction

4.4 Results Comparison

To verify the effectiveness of the proposed method, we compare it with the other four methods in deep learning, U-Net [36], FDU-Net [37], KI-GAN [27], Y-Net [28]. All of them are the latest or the best, or the most basic reconstruction network. These methods can be summarized as follows: 1) U-Net: Generally, an initial inversion is used to reconstruct an image with artifacts from the sensor data. U-Net is then applied as a post-processing step to remove artifacts and improve image quality. 2) FDU-Net: This method employs a fully dense U-Net for removing artifacts from 2-D PAT images. 3) KI-GAN: This method infuses classical signal processing and certified knowledge into the Generative Adversarial Networks for PA imaging reconstruction. 4) Y-Net: This method connects two encoders with one decoder path and integrates both raw data and beamformed images as input to reconstruct the PA image.

We duplicate the above models following the original paper, keep all the original hyperparameters and evaluate the proposed model in both simulated and *in vivo* data. Fig. 6 and Fig. 8 show qualitative comparisons between the proposed method and others on simulated and *in vivo* data, respectively. For sparse data, it is hard for DAS to reconstruct the texture information of the objects, while deep learning networks can recover the texture information better. Our approach can even reduce artifacts in the ground truth, which is related to raw data as input and attention mechanism in the AS-Net. Specifically, in this paper, the ground truth is obtained by DAS reconstruction using dense PA signals. Artifacts will be generated during the DAS, but the original PA signal does not contain these artifacts. In addition, the application of the attention mechanism in the SFE module can make the network pay more attention to semantic features so that some artifacts produced in the DAS process can be filtered out.

Table I shows the mean and standard deviation of SSIM and PSNR of these methods. The AS-Net outperforms other methods in SSIM. For PSNR, AS-Net performs better than others in Mice Data *in vivo*, and FDU-Net is better for Simulated Vessel Data and Fish Data *in vivo*. From Fig. 6 and Fig. 8, we found fewer artifacts in AS-Net's results than the ground truth, and PSNR pays more attention to pixel-level differences. These two points can explain why the performance of PSNR is not the best. In [38], we explain why the output result is better than the ground truth. Considering the residual structure is employed in AS-Net, the PSNR value could be lower than others. Besides, due to the high instability and hyperparameter sensitivity of GAN, it is difficult to adapt to different datasets, which makes KI-GAN perform poorly on our dataset.

## 4.5 Complexity comparison

As mentioned above, the proposed PA data processing method, folded transformation, can reduce model parameters and speed up the PA image reconstruction. In order to justify the statement, we conducted a comparative experiment, which includes two groups with and without the proposed folded transformation. Then, we calculated the model size, floating-point per second, and time to test a sample input. As shown in Table II, for our method with the proposed folded transformation, the model size is decreased by 62% from 27.923M to 10.63M, and GFLOPs are reduced by 33.4% 90.725G to 60.396G. The proposed data preprocessing method can speed up the model running time by 21%. Results indicate that folded transformation can significantly reduce the space and time complexity of the reconstructed model.

## 4.6 Ablation study

To further validate the validity of the proposed method, a series of ablation experiments were set up on the Simulated Vessel Dataset.

TABLE III
THE QUANTITATIVE RESULTS OF ABLATION STUDY (MEAN±STD).

| | U-Net w/o FT | U-Net w/ FT | AS-Net w/o FT | AS-Net w/o SFE | AS-Net w/o BPR | AS-Net w/o $L_{aux}$ | AS-Net |
|---|---|---|---|---|---|---|---|
| SSIM | 0.3458±0.0357 | 0.4595±0.0667 | 0.8531±0.0342 | 0.6276±0.0444 | 0.8565±0.0272 | 0.8616±0.0382 | **0.8953±0.0311** |
| PSNR | 11.6057±0.8405 | 12.0959±0.9827 | 18.8699±1.5662 | 14.0360±1.7607 | 19.1229±1.7664 | 18.8854±1.3791 | **19.5235±1.7286** |

*1) Verification for folded transformation*

The folded transform was first proposed as a very ingenious method for PA signal processing. Its effect on reducing the model complexity has been verified in previous chapters (Table II). To further verify its effect for PA reconstruction, we set up the following experiments: a). our AS-Net with the folded signal as input (our method); b). AS-Net with the original signal input (without FT). Since the original signal is not a square matrix, the final output requires to be a square image. The bottom convolution of the BPR module in AS-Net is set to 20×3 kennel size and 20 step size [27]; c). U-Net with the folded signal as input; d). U-Net with the original PA signal as input, Also, the bottom convolution of U-Net is set to kennel size 20×3 and step size 20. The experimental results are shown in Table III.

The experimental results illustrate that both the proposed AS-Net and the basic U-Net model achieve better results with the folded transformed signal input.

*2) Verification of the combined input of the original PA signal and DAS image*

Since DAS images are simple and easy to obtain, they are employed as auxiliary inputs in this work to provide semantic features. In order to verify the contribution of PA signal and DAS image as input to the reconstruction, a set of comparison experiments are established: only the BPR module is kept for reconstruction, and only the SFE module is kept. Their reconstruction results are compared with the entire AS-Net. Table III shows the reconstruction results.

The experimental results show that the reconstruction results of the SFE module with only DAS image input are better than those of the BPR module with only PA signal input. However, the reconstruction results of both are inferior to those of the AS-Net with combined PA signal and DAS image as input.

*3) Verification of the auxiliary loss Laux at the SFE module*

$L_{aux}$ is aimed at supervising the SFE module to better extract the semantic features while reducing the pressure of L recon. Both of the loss functions promote each other to improve the quality of the final reconstruction. A group of comparison experiments with and without $L_{aux}$ also verify this conclusion. The result is shown in Table III.

The experimental results show that the reconstruction results of the AS-Net with $L_{aux}$ removed decreased by 3.8% in SSIM and 3.2% in PSNR, which indicates that $L_{aux}$ plays a non-negligible positive role in our method.

# 5. Discussion

In this paper, we achieved the goal of reconstructing images from sparse PA signals. A good reconstruction algorithm of sparse signals can help perform photoacoustic imaging with fewer probes and reduce the imaging cost significantly.

For the processing of unsquared PA signals, we proposed folded transform method. The PA signal is transformed into a multi-channel square matrix without information missing and preserving its spatial structure as much as possible. Folded transformation can avoid using large convolution kernels and large step sizes in the neural network to turn the unsquared signals into images [27], and significantly reduce the network parameters and speed up the reconstruction. Experimental results show that the folded transform improves the reconstruction performance to a certain extent.

In this work, both the original PA signal and the DAS image are entered into the network. The DAS image is simple, accessible and contains the basic image features of the subject, but the DAS image includes numerous artifacts. The original PA signal involves the complete imaging information, but turning the PA signal into an image is complicated. This work extracts as many features as possible from the PA signal and the DAS image to facilitate reconstruction.

# 6. Conclusion

PA image reconstruction is essential in photoacoustic imaging. In this paper, a novel PA signal preprocessing method is employed to accelerate the reconstruction. Furthermore, an end-to-end deep learning framework named AS-Net is proposed to reconstruct the PA image from sparse PA data and DAS images. A global context block is utilized to preserve more spatial information. Our experimental results show that the proposed method can improve the accuracy and speed of the PA image reconstruction in different tasks, including simulated fundus vessel PA image reconstruction, fish *in vivo* PA reconstruction, mice *in vivo* PA reconstruction. In this paper, the method is validated using a time-domain signal, and the extension to the frequency domain signal would be a possible work in the future.